\documentclass[a4paper, conference]{IEEEtran}
\IEEEoverridecommandlockouts
\usepackage{cite}
\usepackage{amsmath,amssymb,amsfonts}
\usepackage{algorithmic}
\usepackage{graphicx}
\usepackage{textcomp}
\usepackage{multirow}
\usepackage{hhline}
\usepackage{xcolor}
\usepackage{hyperref}
\usepackage{multirow}
\usepackage{rotating}
\usepackage{hhline}
\usepackage[caption=false, font=footnotesize]{subfig}
\begin{document}

\title{SyNet: An Ensemble Network for Object Detection  \\in UAV Images}

\author{\IEEEauthorblockN{Berat Mert Albaba}
\IEEEauthorblockA{\textit{Dept. of Mechanical Engineering} \\
\textit{Bilkent University}\\
Ankara, Turkey \\
mert.albaba@bilkent.edu.tr}
\and
\IEEEauthorblockN{Sedat Ozer}
\IEEEauthorblockA{\textit{Dept. of Computer Engineering} \\
\textit{Bilkent University}\\
Ankara, Turkey \\
sedat@cs.bilkent.edu.tr}
}

\maketitle

\begin{abstract}

Recent advances in camera equipped drone applications and their widespread use increased the demand on vision based object detection algorithms for aerial images. Object detection process is inherently a challenging task as a generic computer vision problem, however, since the use of object detection algorithms on UAVs (or on drones) is relatively a new area, it remains as a more challenging problem to detect objects in aerial images. There are several reasons for that including: (i) the lack of large drone datasets including large object variance, (ii) the large orientation and scale variance in drone images when compared to the ground images, and (iii) the difference in texture and shape features between the ground and the aerial images.  
Deep learning based object detection algorithms can be classified under two main categories: (a) single-stage detectors and (b) multi-stage detectors. Both single-stage and multi-stage solutions have their advantages and disadvantages over each other. However, a technique to combine the good sides of each of those  solutions could yield even a stronger solution  than each of those solutions individually. 
In this paper\footnote{This paper is the preprint (ArXiv) version of its accepted ICPR2020 version. }, we propose an ensemble network, SyNet, that combines a multi-stage method with a single-stage one with the motivation of decreasing the high false negative rate of multi-stage detectors and increasing the quality of the single-stage detector proposals. As building blocks, CenterNet and Cascade R-CNN with pretrained feature extractors are utilized along with an ensembling strategy. We report the state of the art results obtained by our proposed solution on two different datasets: namely MS-COCO and visDrone with \%52.1 $mAP_{IoU = 0.75}$ is obtained on MS-COCO $val2017$ dataset and \%26.2 $mAP_{IoU = 0.75}$ is obtained on VisDrone $test-set$. Our code is available at: \href{https://github.com/mertalbaba/SyNet}{https://github.com/mertalbaba/SyNet}
\end{abstract} 


\section{Introduction}

Object detection is an essential yet a challenging task in computer vision field, despite the recent advances using deep learning based techniques. While object detection demonstrated significant success in many applications, those applications are mostly limited in domains using ground taken images. \textit{ImageNet}~\cite{deng2009imagenet}, \textit{COCO} \cite{Microsoft2015coco} and \textit{PASCAL}~\cite{everingham2010pascal} datasets are such major datasets driving the deep learning based object detection algorithms and they contain images taken predominantly on the ground. Consequently, they do not include a large variance of the object properties as observed from air. Furthermore, such state of the art object detection algorithms are designed considering the problems associated with the ground images mainly. That is why even the most recent works focusing on aerial images such as \cite{wu2019delving}, \cite{hong2019SungeunICCV}, \cite{RRNet} cannot reach to the level where the state of the art object detection algorithms perform today on ground images. With the motivation behind deep neural networks and how object detection is being performed, performance of the state-of-the-art object detectors suffer mainly from two reasons in challenging settings: (1) the images captured has a non-uniform distribution in terms of object types and (2) scale, orientation, shape and the features of the objects can significantly differ from the ones appearing in the ground images.  

In recent studies considering class imbalance, most studies addressed the class imbalance problem from considering the imbalance between the foreground and background of the image samples \cite{zhu2020vision}, \cite{Hong2019patch}. As another form of class imbalance, the imbalance between the foreground objects also effects detection performance in majority of algorithms, as some of those relevant issues were addressed with different data augmentation techniques as in \cite{Hong2019patch}. Considering the aerial image setting and the existing image samples in \textit{VisDrone} dataset \cite{zhu2020vision} (which we also use in this paper), class imbalance between the foreground objects remains as a common issue that effects the prediction performance negatively. The problem of class imbalance also shows its significance in ground-level view, which can be observed in \textit{MS-COCO} dataset \cite{Microsoft2015coco}. As also discussed in \cite{Yang_2019_ICCV}, scale of the objects in datasets is another issue effecting the detection performance. 

Typically, each existing object detection algorithm has its advantages and disadvantages over the other existing detection algorithms. Therefore, combining the advantageous properties of multiple detection algorithms can yield a higher performance and that is what we mainly focus on and introduce in this paper for aerial images. We propose an ensemble network that selects the good properties (good decisions) of two recent state of the art detection algorithms namely Cascade R-CNN \cite{cai2019cascadercnn} and CenterNet \cite{Duan2019centernet} to achieve even better state of the art results on aerial images. In order to deal with small objects for which assessing detection performance using Intersection over Union (IoU) metric would not be trivial, benefits of the Cascade R-CNN \cite{cai2019cascadercnn} algorithm are utilized in our proposed solution. In addition to that, recently proposed CenterNet \cite{Duan2019centernet} is also used in our proposed solution, as CenterNet is another effective detection algorithm with good detection rates at different scales. CenterNet is oriented at the center of the detected objects and that enables the method to be more robust to scale differences. %give accurate results for objects that are scaled different from the ideal setting. 

%that yields the state of the art results on \textit{VisDrone} dataset

% In this work, in order to deal with small objects for which assessing detection performance using Intersection over Union (IoU) metric would not be trivial, benefits of the method Cascade R-CNN \cite{cai2019cascadercnn} are utilized. In addition to that, recently proposed CenterNet \cite{Duan2019centernet} is also used in our proposed solution, as CenterNet is also effective in detecting objects at different scales. CenterNet is oriented at the center of the detected objects and that enables the method to give accurate results for objects that are scaled different from the ideal setting. 

Our proposed approach includes two steps for detection: (i) we employ an image augmentation technique inspired from \cite{dwibedi2017cut} to deal with the imbalance problem making our proposed approach more robust to classes with lower number of samples. (ii) we introduce a synergistic network which combines the predictions of multiple detectors and chooses the most confident score through a weighted bounding box fusion \cite{solovyev2019weighted} method to detect objects. Combining those two steps offers state of the art prediction performances, when compared to previous recent efforts in the field \cite{renNIPS15fasterrcnn}, \cite{he2017mask}, \cite{Duan2019centernet}, \cite{cai2019cascadercnn} by considering and dealing with two important problems: the class imbalance problem and the scaling problem. 

As verified by our experiments, our proposed method SyNet achieves state of the art performance in object detection, when compared to the performances of the state-of-the-art methods on the VisDrone \cite{zhu2020vision} dataset. In addition to the aerial images, we also observed that proposed approach works well and achieves higher performance for object detection in ground-based images, as our experiments with \textit{MS-COCO} dataset \cite{lin2014microsoft} demonstrates. Overall, our main contributions can be listed as:
\begin{itemize}
    \item introduction of a synergistic approach for better object detection that combines the advantageous properties of two individual and state of the art object detectors, 
    \item demonstrating the performance of our proposed object detection method not only on a drone dataset: \textit{VisDrone} but also on a ground-image based dataset: \textit{MS-COCO},
    \item employing an augmentation technique to deal with class imbalance problem in aerial images. 
\end{itemize}

% All in all, our approach mainly focuses on using the double-stage detector Cascade R-CNN\cite{cai2019cascadercnn} as the baseline for object detection tasks yet it benefits from the single-stage architecture CenterNet \cite{Duan2019centernet}. Integrated data augmentation before the detectors yields better results for the detection task. 

\section{Related Work}
Object detection has received significant amount of attention in the last decade. In this section, most relevant work to ours are summarized under two subcategories: (i) Object Detection and (ii) Image (data) Augmentation.

\subsection{Object Detection}
With the recent developments in deep learning, great advancements are achieved in multi-class image recognition. With ImageNet \cite{imagenet_cvpr09} and Pascal VOC \cite{everingham2010pascal} challenges, with utilization of convolutional neural network (CNN) blocks along with additional functions such as max-pooling, successful methods are proposed for image recognition. In 2012, a very deep CNN is proposed for image recognition in \cite{krizhevsky2012imagenet}, which results a major improvement in image recognition. Next, rectified linear units \cite{agarap2018deep} are introduced in \cite{zeiler2014visualizing} and in \cite{simonyan2014very}, it is showed that simpler CNN with less filters can also be successful for image recognition. In 2014, skip connections are introduced into convolutional blocks, inception modules, in \cite{szegedy2015going}. Then, a very deep network with skip connections \cite{he2016deep} has achieved great performance.  

Following the success of deep convolutional neural networks (CNNs) in image classification \cite{krizhevsky2012imagenet}, CNNs are also used for a more challenging task: object recognition and localization, i.e. object detection. Firstly, single forward-pass networks are proposed for object recognition, which are referred as singe-stage detectors. In \cite{redmon2016you}, object detection is considered as a regression problem. After dividing image into grids containing pixels, bounding box and class predictions are made for each cell of the grid and good performance is seen in \cite{pascal-voc-2012}. In \cite{redmon2017yolo9000}, batch normalizations and convolutions with anchor boxes are introduced into \cite{redmon2016you}. Then, in \cite{liu2016ssd}, anchor boxes, fixed size bounding boxes which are used as reference boxes, are utilized along with multi-task loss for object detection and a great performance is achieved in different extensive datasets. In \cite{fu2017dssd}, ResNet101 is introduced into SSD and deconvolution layers are used to obtain better detection method, which can perform better for small-scale objects. In \cite{lin2017focal}, a new loss called $Focal Loss$ is utilized to solve the class imbalance problem in the object detection. Then, in \cite{zhang2018single}, a module that refines anchor boxes are proposed. In \cite{law2018cornernet}, anchor boxes are fully eliminated, objects are detected as two keypoints defining the bounding box and corner pooling is introduced. In \cite{farhadi2018yolov3}, for each class, logistic classifiers are utilized and better feature extractor is used when compared to previous version \cite{redmon2017yolo9000}. Recently, in \cite{Duan2019centernet}, objects are detected as only a single keypoint, that contains the center location of the object, and other properties such as size are regressed, which results good performance with less computational complexity. 

In contrast to single-stage detectors, there are also multi-stage detectors, i.e. double-stage detectors, which divides detection problem into two sub-problems: 1. region proposals and 2. bounding box regression. In \cite{girshick2014rich}, nearly 2000 region proposals are generated through selective search and after that, support vector machines and CNN are used to regress bounding boxes and class labels. Then, in \cite{girshick2015fast}, region of interest pooling is introduced into \cite{girshick2014rich} and instead of finding regions directly on the image, regions are generated from the feature vector of the image. In \cite{ren2015faster}, for the region proposals, a new approach is presented: Region Proposal Network (RPN), which is a CNN for producing region proposals and seperately trainable. Several reference boxes are utilized in \cite{ren2015faster}. Region proposals generated by RPN are fed into the region of interest pooling layer and a detection network generates bounding box and class label predictions from pooled region proposals. In \cite{dai2016r}, position sensitive region of interest pooling layers are proposed and region proposals are generated from the full image, instead of from several sub-regions. In \cite{he2017mask}, region of interest alignment is introduced into \cite{ren2015faster} and better results are achieved. 

In addition to single-stage and multi-stage detectors, there also exists some cascading approaches utilizing sequentially cascaded networks, such as \cite{PANet} and \cite{cai2019cascadercnn}. In \cite{cai2019cascadercnn}, sequential detectors whose quality levels increases with the increase in the depth, i.e. stage number, is proposed and great improvements are seen in state-of-the-art double-stage detectors which are explained previously.

\subsection{Data Augmentation}
To solve the class imbalance problem, data augmentation is a popular technique that is used by applying fundamental operations to images such as rotation, scaling and flipping. Moreover, augmentation methods provide an enrichment on the data that the technique is used. In addition to traditional augmentation techniques like flipping, rotating, translating or coloring, different data augmentation methods are proposed.  

\begin{figure*}
	\centering
	\includegraphics[width=\textwidth]{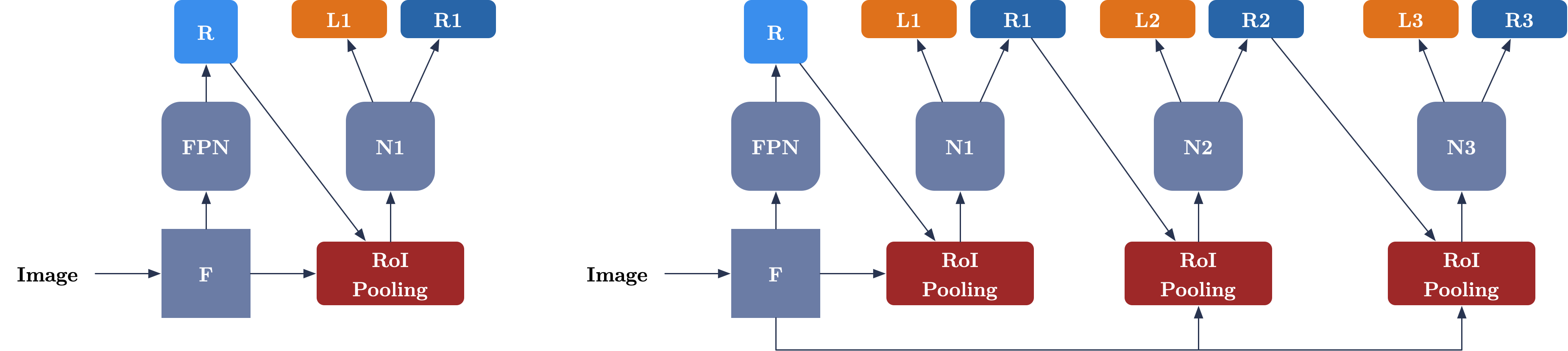}
	\caption{Architectures of the Faster R-CNN (on the left) and Cascade R-CNN (on the right). $F$ is the feature extractor, $N_i$s are network bodies, $R$ is the region proposals, $L_i$s are label predictions and $R_i$s are the bounding box predictions.}
	\label{casc}
\end{figure*} 

In \cite{dwibedi2017cut}, objects which should be detected by object detectors are cutted, i.e. copied, and pasted on several background images, along with altering pixel artifacts to generate new training samples. A technique that randomly removes some parts of the images to create new training samples with occlusions is proposed in \cite{zhong2017random}. In \cite{inoue2018data}, a overlaying images method named $SamplePairing$ is proposed which mixes two different images and takes average values of pixels. In \cite{liang2018understanding}, generative adversarial neural networks (GANs) are utilized to produce new training samples by mixing different images. In \cite{tanaka2019data}, GANs are utilized for data augmentation and it is showed that the generated data may be more beneficial than the original ones. In \cite{zhan2019spatial}, spatial fusion generative adversarial neural networks (SF-GANs) are used for new image generation from original ones. In order to overcome the problems of distribution in aerial images, a data augmentation approach is proposed by \cite{Hong2019patch}, which resulted with significant increase in prediction performance in terms of mAP score. Lastly, in \cite{RRNet}  as an augmentation technique, AdaResampling resolves the problems of scale and background mismatch.

\section{ Methods}
For object detection, different detectors with different architectures, some of which have multiple steps, are utilized in the literature (as discussed in the previous section). In this work, a method that combines single-stage and multi-stage object detection architectures synergistically is proposed. Furthermore, image augmentation is utilized in order to enhance the results. Therefore, next, we describe the details of the used methods in our proposed solution. 

\subsection{Object Detection}
In this paper, a neural network based object detector is proposed that aims to localize and recognize several objects on given images. A detector computes a bounding box, which, typically, is a quadruple $(x^p_{tl}, y^p_{tl}, x^p_{br}, y^p_{br})$ where $x^p_{tl}$, $y^p_{tl}$ are the coordinates of the top-left corner of the prediction, $p$, and $x^p_{br}$, $y^p_{br}$ are the coordinates of the bottom right corner. For evaluating the quality of the prediction, as a common metric, intersection of union ($IoU$) is utilized, which can be defined as $IoU(b^p, b^{gt}) = \dfrac{a(b^p\cap b^{gt})}{a(b^{p} \cup b^{gt})}$, where $a$ is the area function that calculates the area of the given bounding box, $(x_{tl},y_{tl},x_{br},y_{br})$, $b^{p}$ is the predicted bounding box and $b^{gt}$ is the true bounding box. Fig. \ref{iou} shows the general process.

For the evaluation of the prediction based on the $IoU$ metric, a threshold $t$ is selected and if the $IoU(b^{p}, b^{gt})$ score is larger than the threshold, prediction id considered as the successful. Usually, prediction threshold $t$ is referred as the quality level of the detection network.

In our proposed solution, two individual detectors are utilized: Cascade R-CNN and CenterNet. Next, we describe each of those detectors. 

\subsubsection{Cascade R-CNN}
Cascade R-CNN is a cascaded network, that contains multiple repeated networks, which are connected sequentially, and based on the Faster R-CNN \cite{renNIPS15fasterrcnn} method. In other words, Cascade R-CNN is the sequentially repeated version of the Faster R-CNN. In this work, Cascade R-CNN is utilized as the first object detection building block. Detailed explanation of Cascade R-CNN can be found in \cite{cai2019cascadercnn} and this method is briefly explained here.

For object detection, there are two main problems: (1) a detection network optimized for a certain intersection of union (IoU) score $q$, may not perform well in the test dataset, since the network is optimized for $q$ and IoU scores of the test samples may significantly differ from $q$; and (2) number of samples decreases seriously with the increase in the overall IoU scores of the samples and this makes a high quality (IoU score) detector prone to over-fitting problem since the number of samples are not enough. To overcome these problems, a cascaded network is proposed to increase the number of high IoU score samples and allow the detector to perform well for different IoU scores, whose architecture is presented in Fig. \ref{casc}. 

For the bounding-box predictions, a single detector, $d$, tries to minimize bounding-box regression loss \cite{cai2019cascadercnn}, which is defined as
\vspace{-0.1in}
\begin{equation}
L(d) = \sum_{i} G(b^p_i,b^{gt}_i)
\end{equation} 

\noindent where
\vspace{-0.1in}
\begin{equation}
\begin{split}
G(b^1,b^2) = H(x^1_{tl}-x^2_{tl}) + H(y^1_{tl}-y^2_{tl}) + \\ H(x^1_{br}-x^2_{br}) + H(y^1_{br}-y^2_{br})
\end{split}
\end{equation} 

\noindent and

\vspace{-0.15in}
\begin{equation}
H = \begin{cases}
      0.5 x^2, & \text{if}\ |x| < 1 \\
      |x|-0.5, & \text{otherwise}
    \end{cases}
\end{equation}

\noindent in which $b_i^p$ is the bounding box prediction of $i^{th}$ object, $b_i^{gt}$ is the true bounding box of $i^{th}$ object, and $(x_{tl},y_{tl})$ and $(x_{{br}},y_{{br}})$ are the top-left corner coordinates and  the bottom-right coordinates of a bounding box, respectively. In cascaded structure, as shown in Fig. \ref{casc}, $N$ bounding box regressors are utilized instead of a single regressor and the final bounding box predictions are generated as

\begin{equation}
b^p = d_N \circ d_{N-1} \circ \ldots \circ d_1(x,b^{gt})
\end{equation}

\noindent where $x$ is the input image and $\circ$ is an iterative operator such that each regressor uses the outputs of the previous regressor. As showed in \cite{cai2019cascadercnn}, this allows the network to perform well in different IoU scores of inputs since each regressor are trained with different sample qualities and optimized for different IoU scores. Furthermore, in Cascade R-CNN, a detector $d_n$ at stage $n$ minimizes:

\begin{equation}
L(x_n, b^{gt}) = L_C(c_n(x_n),y_n)+ I[y_n \geq 1]L_L(d_n(x_n,b_n^{p}), b^{gt})
\end{equation}

\noindent where $c_n$ is the classifier at stage $n$, $b_n^{p} = d_{n-1}(x_{n-1},b^p_{n-1})$, $x_n$ is the input to stage $n$, $y_n$ is the label of $x_n$, $L_C(c_n(x_n),y_n)$ is classification loss defined as
\begin{equation}
L_C(c_n(x_n),y_n) = -log(p(y=y_n|x))
\end{equation}
and $L_L(d_n(x_n,b_n^{p}), b^{gt})$ is defined as
\begin{equation}
L_L(d_n(x_n,b_n^{p}), b^{gt}) =  G(d_n(x_n,b_n^{p}), b^{gt}) .
\end{equation}
In this structure, since the outputs of the previous detector $d_{n-1}$, which is optimized for IoU score $1_{n-1}$ are fed to stage $n$, which is optimized for $q_n$ such that $q_n > q_{n-1}$, IoU score of the generated samples are gradually increased. In other words, as going deeper in the cascaded structures, number of high IoU score samples increases, which provides enough samples to higher level detectors and prevent over-fitting issue. To clarify, IoU score level increases as going deeper in the cascaded structure since the IoU score of inputs increases.

As a summary, by resampling the initial samples, the IoU scores of the samples are increased sequentially and a network robust to over-fitting is obtained. Moreover, since detectors at different stages are optimized for different IoU scores, mismatches between training and test performances is eliminated at a higher level. Hence, Cascade R-CNN offers high quality detection results. 

\subsubsection{CenterNet}
\begin{figure}[!b] 
\vspace{-0.3in}
    \centering
  \subfloat[]{%
       \includegraphics[width=0.45\linewidth]{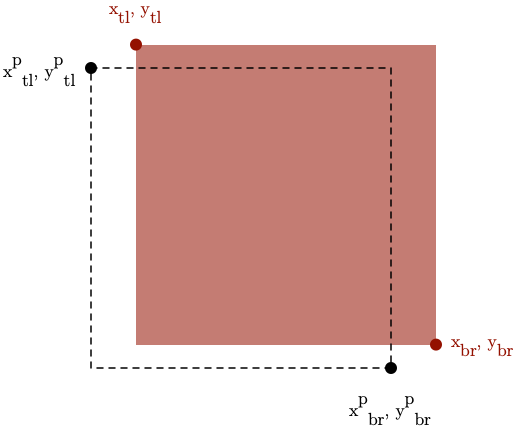}}
    \hfill
  \subfloat[]{%
        \includegraphics[width=0.45\linewidth]{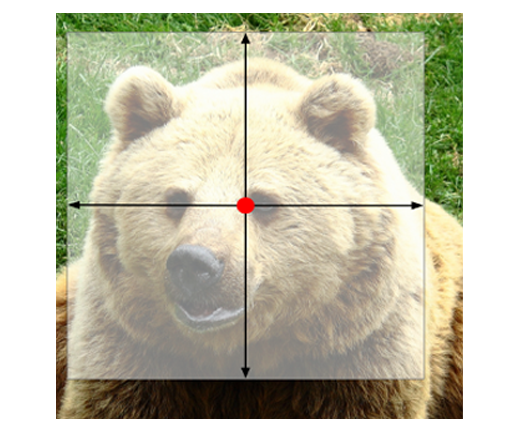}}
        \vspace{-0.05in}
  \caption{(a) The object (red box) and the predicted bounding box (dashed box). (b) Estimated center point and sizes of the object by CenterNet.}
  \label{iou} 
\end{figure}

In most of the object detection architectures, objects are detected by utilizing some anchors, which are fixed size bounding boxes. In contrast, as shown in Fig. \ref{iou}, CenterNet \cite{zhou2019objects} approaches object detection as a keypoint estimation problem, which allows more flexible, faster object detection.

\begin{figure*}
	\centering
	\includegraphics[width=\textwidth]{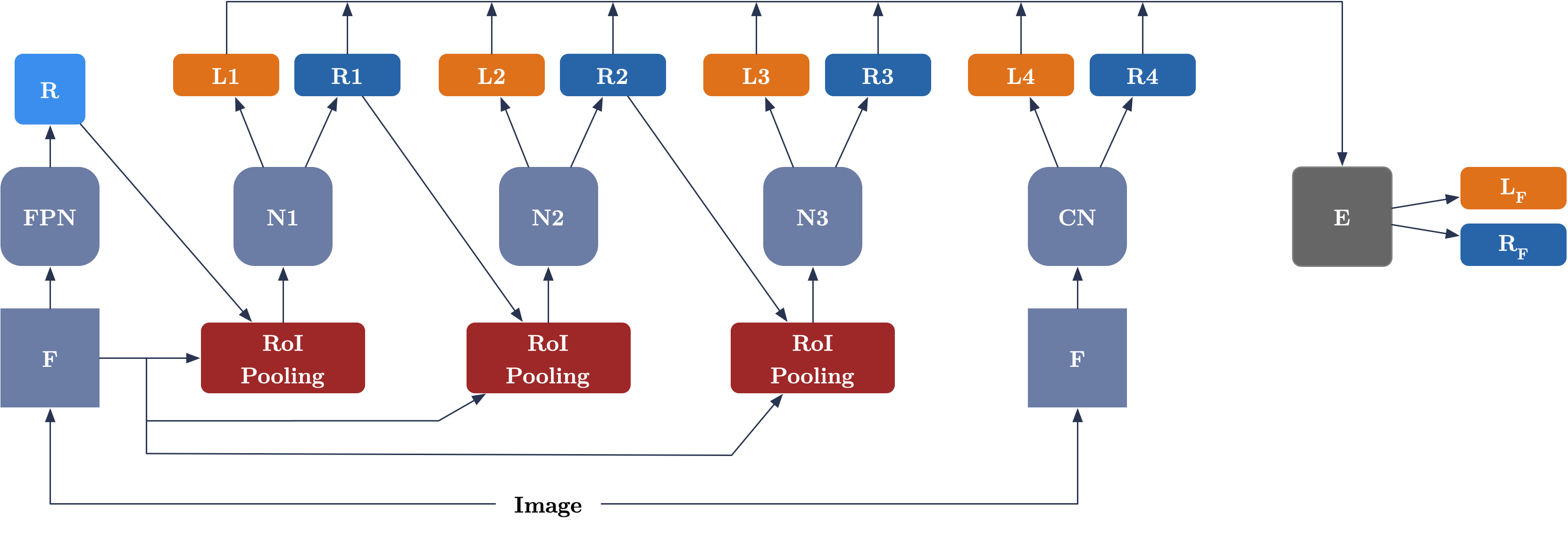}
	\vspace{-0.4in}
	\caption{Architecture of  SyNet. $N_i$s are network bodies containing convolutional layers, $F$ is the feature extractor, $CN$ is the CenterNet body, $R$ is the region proposals, $L_i$s are label predictions, $R_i$s are bounding box predictions, $E$ is the ensemble function and $L_F, R_F$ are final label and bounding box predictions, respectively.}
	\label{comb}
\end{figure*}

In CenterNet approach, first of all, a heatmap of keypoints is predicted as $Y^p$, in which 1 represents a keypoint detection and 0 is background label. For a true keypoint tuple $gt_i = (x^{gt}_i, y^{gt}_i) \epsilon R^2$, a heatmap $Y^{gt}_i$ is created as

\begin{equation}
Y^{gt}_i = e^{\dfrac{-(x^{gt}_i-k^{quant_i}_x)^2-(y^{gt}_i-k^{quant_i}_y)^2}{2\sigma^2_{k^{quant_i}}}}
\end{equation}

\noindent where $k^{quant_i} = \lfloor x^{gt}_i/{R},y^{gt}_i/{R}\rfloor$, where $R$ is the output stride, and $\sigma$ is standard deviation which is size adaptive.  Then, as explained in \cite{zhou2019objects}, a network that predicts keypoints is trained by minimizing the loss defined as

\begin{equation}
\begin{split}
L = \dfrac{-1}{K} \sum_{gt_i} g[Y^{gt}_i](1-Y^p_{gt_i})^2 log(Y_{gt_i}^p) +\\ g[Y^{gt}_i](1-Y^{gt}_i)^4(Y^p_{gt_i})^2 log(1-Y_{gt_i}^p)
\end{split}
\end{equation}

\noindent where $g(x)=x$ if $x=1$ and $g(x) = 1-x$ otherwise and $K$ is number of keypoints. In addition to keypoints, an offset, $O_{gt_i}$ is predicted for each point in order to recover error from discretization by minimizing $L_o= \dfrac{1}{K}\sum_{gt_i}|O_{k^{quant_i}}+k^{quant_i}-gt_i/R|$. 

After keypoint predicting network is trained, for each object, keypoints are utilized as center points of the objects, which have a bounding box represented as quadruple $(x_{lt}, y_{lt}, x_{br}, y_{br})$. For each object classes, $c$, width and height predictions are made as $v^p_c = (w^p_c, h^p_c)$ to predict true width and height of $i^{th}$ object from class $c$, $(v^gt_c)_i = (x^i_{lt}-x^i_{br}, y^i_{lt} - y^i_{br})$ by minimizing

\begin{equation}
L_v = \dfrac{1}{n}\sum_i |v^p_c-(v^gt_c)_i|
\end{equation}

\noindent for each class.  Then, a single loss is defined as weighted sum of $L$, $L_v$ and $L_o$, which is used in a single network for all predictions \cite{zhou2019objects}. For bounding box predictions, detected local maximum keypoints are used as center points for each class and a bounding box quadruple is predicted as $(x^i_{lt}+o^i-w^i/2, y^i_{lt}+q^i-h^i/2, x^i_{br}+o^i+w^i/2, y^i_{br}+q^i+h^i/2)$, where $(o,q)$ is the offset predictions and $(w,h)$ is the size predictions. 
As result, a powerful and fast object detector is obtained which does not require any post-processing like bounding box suppressions and any fixed anchor boxes. 

\subsubsection{Weighted Bounding Box Fusion}
For the combination of different methods, an ensemble method named weighted boxes fusion approach \cite{solovyev2019weighted} is used. As explained in \cite{solovyev2019weighted}, instead of removing some bounding box predictions like non-maximum suppression, this method utilizes all predictions in order to find bounding box clusters.

Initially, clustered boxes are stored in $C$. $C_i$ contains the matched boxes whose labels are $i$. Iteratively, each bounding box prediction with $j^{th}$ label is compared with $F_j$, which is the fusion of boxes in $C_j$ and represented by quintuple $(s, x_{tl}, x_{br}, y_{tl}, y_{br})$, where $s$ is the confidence score. When a box $b$ is matched with $F_j$ and added to $C_j$, fusion box $F_j$ confidence is recalculated as

\begin{equation}
F_j(s) = \dfrac{\sum_{A_k \epsilon C_j} A_k(s)}{N}
\end{equation}

\noindent where $N$ represents the total number of boxes in $C_j$ and $A_k$ is bounding box sample, i.e. a quintuple $(s, x_{tl}, x_{br}, y_{tl}, y_{br})$. Coordinates of the fusion box are updated as follows: 

\begin{equation}
F_j(x_1) = \dfrac{\sum_{A_k \epsilon C_j} A_k(s)A_k(x_{tl})}{s_1 + s_2 + \ldots + s_N}
\end{equation}

\begin{equation}
F_j(x_2) = \dfrac{\sum_{A_k \epsilon C_j} A_k(s)A_k(x_{br})}{s_1 + s_2 + \ldots + s_N}
\end{equation}

\begin{equation}
F_j(y_1) = \dfrac{\sum_{A_k \epsilon C_j} A_k(s)A_k(y_{tl})}{s_1 + s_2 + \ldots + s_N}
\end{equation}

\begin{equation}
F_j(y_2) = \dfrac{\sum_{A_k \epsilon C_j} A_k(s)A_k(y_{br})}{s_1 + s_2 + \ldots + s_N}
\end{equation}

\noindent where $s_k = A_k(s)$ is the confidence score of $A_k$. As these equations show, confidence scores of each bounding box is utilized as weights in order to obtain meaningful final bounding box for each class. After all the predicted boxes are compared with clusters sequentially, box confidence scores are rescaled as explained in \cite{solovyev2019weighted}, where it is presented that when compared to common methods such as \cite{zhou2017cad} and \cite{bodla2017soft}. 

\subsection{SyNet}
In this work, a synergistic approach is proposed that combines a single stage detector with a multi-stage detector for better object detection. The main motivation is that multi-stage detectors tend to produce more false negatives, which means that multi-stage detectors fails to detect some objects. On the other hand, single-stage detectors generally propose more bounding boxes with less quality. Hence, the combination of these two different architectures may predict more bounding boxes than multi-stage detectors and quality of the single-stage detector predictions may be increased by the multi-stage one. Thus, an approach that combines Cascade R-CNN \cite{cai2019cascadercnn} and CenterNet \cite{zhou2019objects} through weighted box ensemble \cite{solovyev2019weighted} is proposed. 

General architecture of the proposed method is presented in Fig. \ref{comb}, where $Image$ represents the images in training set, $FPN$ is the feature pyramid network \cite{lin2017feature} for initial features, $R$ is the proposal generated by $FPN$,  $F$ is the feature extractor network, $RoI Pooling$ is the region of interest pooling layer as described in \cite{ren2015faster}, $N1,N2,N3$ are the network bodies consisting several convolutional layers, $CN$ is the CenterNet network body, $L1,L2,L3,L4$ are the label predictions, $R1,R2,R3,R4$ defines the bounding box predictions, $E$ is the weighted box ensemble function that ensembles inputs based on weighted bounding box dusion as explianed in the previous subsection and $L_F, R_F$ are the final class and bounding box predictions, respectively.  As feature extractors, ResNet101 \cite{he2016deep} pretrained on ImageNet \cite{imagenet_cvpr09} is used in Cascade R-CNN and DLA \cite{yu2018deep} is utilized in CenterNet.

First of all, images in the train dataset are augmentated by including additional images, which are generated with the procedure explained in Section 3.1. Then, these are fed into both CenterNet and Cascade R-CNN and these networks are trained. For the Cascade R-CNN,  as the feature extractor, ResNet 101 is used and for CenterNet, DLA is utilized. After the trainings, bounding box predictions of these two networks are obtained for each element in the inference set and these are combined with weighted box fusion method. As result, final bounding box predicitons are obtained. Since the proposed approach requires the training of CenterNet and Cascade R-CNN seperately, memory consumption is significantly higher than both Cascade R-CNN and CenterNet. However, time consumption is similar to that of Cascade R-CNN.

Although CenterNet and Cascade R-CNN are utilized as single-stage and multi-stage detectors, other detectors may also be introduced into the network.

\subsection{Image Augmentation}
With the rapid development of object detectors, the need for extensive data with several local and global features is increased extremely. For this purpose, in this work, data augmentation approach presented in \cite{dwibedi2017cut} utilized.

\begin{figure} 
    \centering
  \subfloat[]{%
       \includegraphics[width=0.45\linewidth]{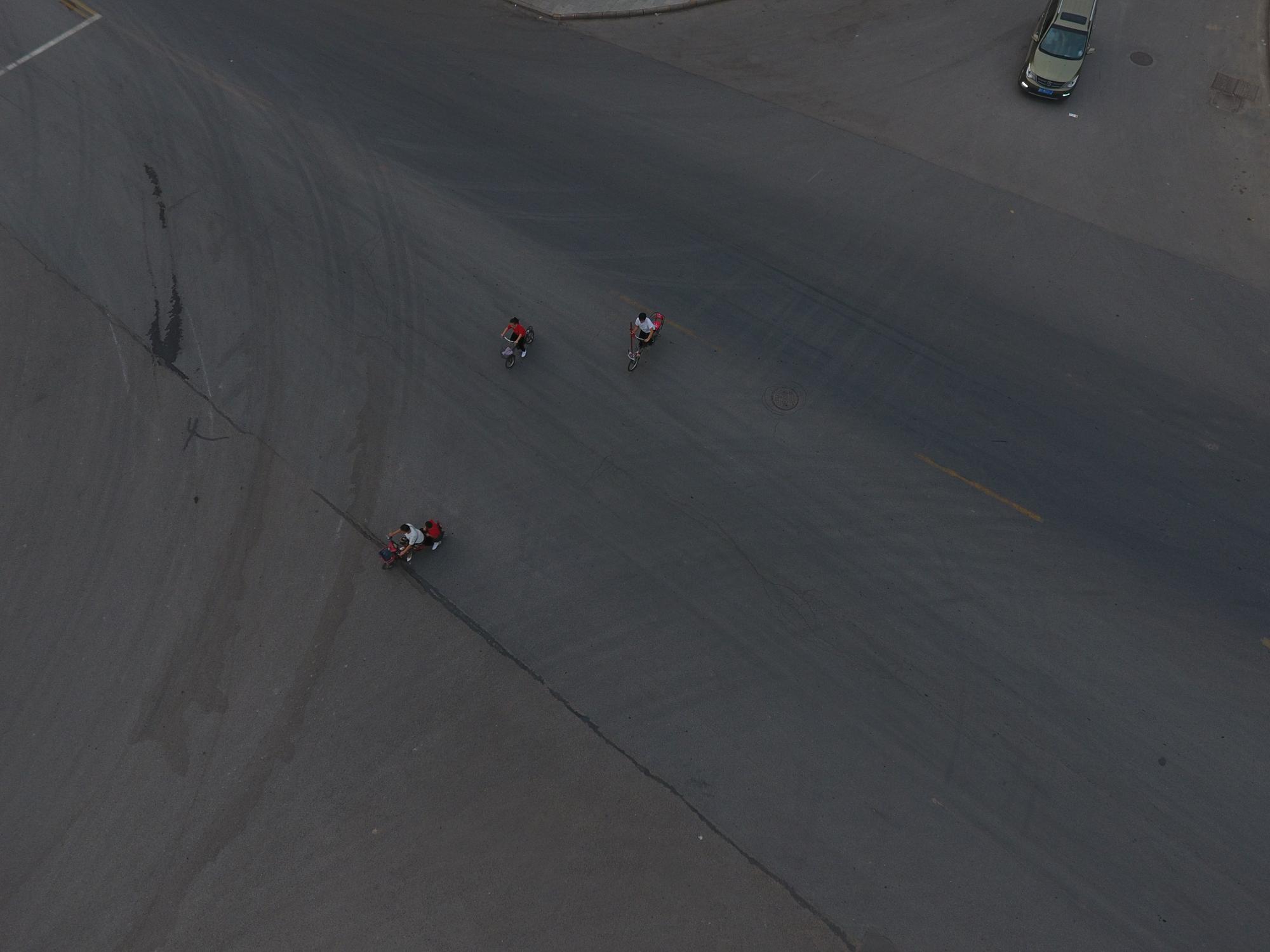}}
    \hfill
  \subfloat[]{%
        \includegraphics[width=0.45\linewidth]{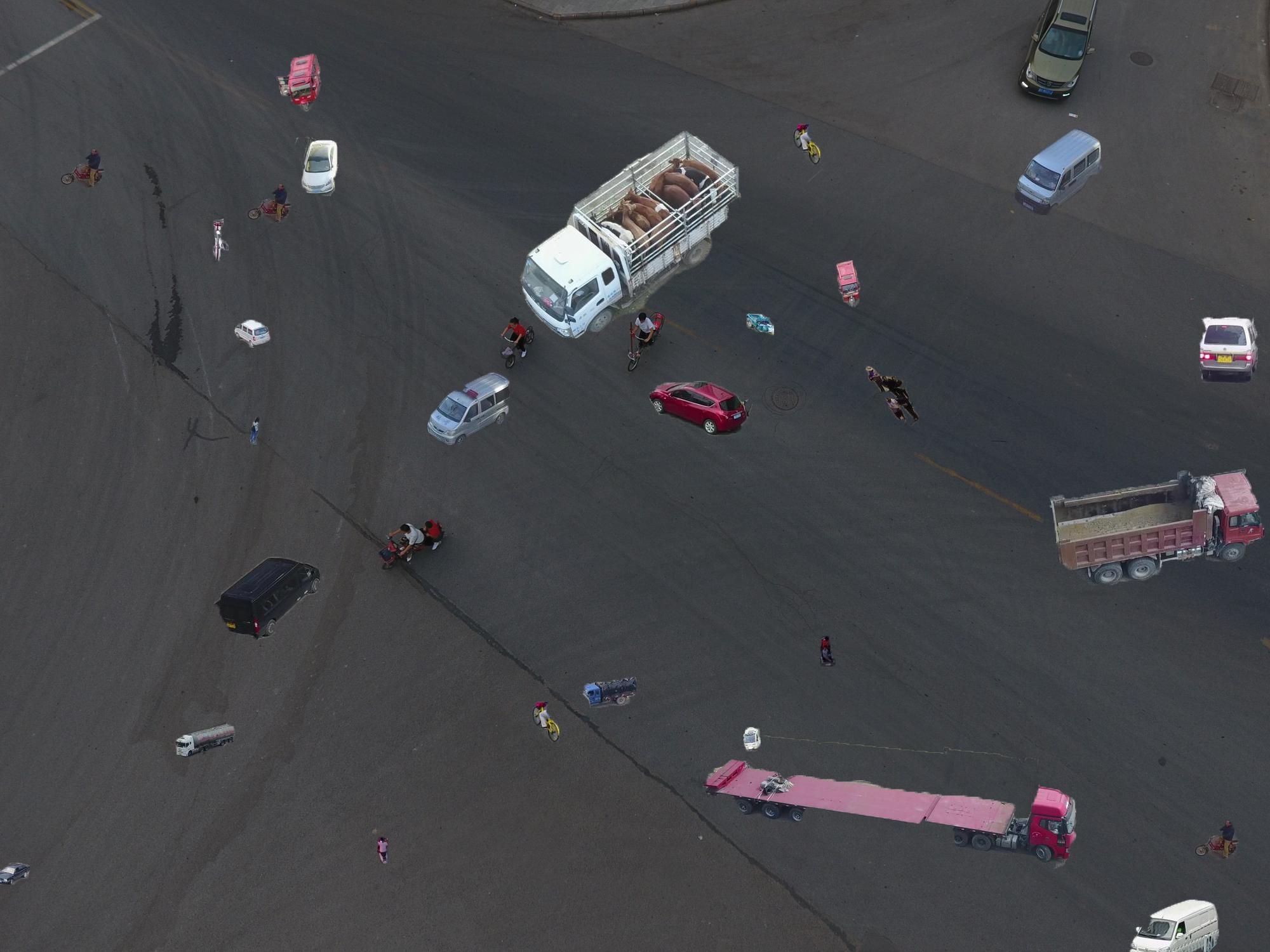}}
  \caption{(a) The original image. (b) The augmented version.}
  \label{augm} 
\end{figure}

In this approach, the main goal is that generating augmented $real-like$ images to allow object detectors work better. As explained in \cite{dwibedi2017cut}, this method consists four main steps. At first, images of objects, which should be predicted by the object detector, are collected. In this work, object images in datasets are utilized. To exemplify, for VisDrone dataset \cite{pailla2019visdrone}, for $car$ class, image parts limited by bounding box coordinates whose labels are $car$ are collected as object instances. After this, scene images should be collected for this approach and in this work, as scene images, raw training dataset images are utilized. Then, masks should be predicted for object images collected at first step. For this purpose, convolutional neural networks are utilized in \cite{dwibedi2017cut}, however, in this work, object segmantations in datasets are used as masks after converting them to binary images if available. Lastly, convolution of masks and objects are placed on randomly choosen scene images. 

In this work, for each image in a dataset, two augmented images, that contains more than 10 and less than 30 pasted additional objects are created to improve training. Fig. \ref{augm} shows an example of the augmentation.

\section{Datasets}
In this work, for the evaluation of the proposed approach, two different datasets are used, one of which is an aerial dataset and the other is a ground dataset.

\subsection{VisDrone Dataset}
VisDrone dataset \cite{pailla2019visdrone} for aerial object detection consists more than 6000 aerial images taken by camera equipped unmanned air vehicles. In total, there are ten classes which are used in the evaluation in both \cite{pailla2019visdrone} and in this work, which may be listed as: 1. pedestrian, 2. people, 3. bicycle, 4. car, 5. van, 6. truck, 7. tricycle, 8. awning-tricycle, 9. bus and 10. motor. For each sample, bounding box coordinates are given along with truncation and occlusion information. In this work, truncation and occlusion information are not utilized since the goal is building a robust detector which may also detect partially truncated or ocluded objects. Since the aerial image detection task is still challenging because of object-image size mismatch and class imbalance, this dataset is utilized in this work for the validation of the proposed approach. 

For the training, training set is utilized that consists 6471 aerial images. Proposed image augmentation method is applied only to the training set. For the validation, for which the results generated, test set is used that consists 1580 images. 

\subsection{MS-COCO Dataset}
For the evaluation of the proposed approach in ground images, Microsoft Common Objects in Context dataset \cite{lin2014microsoft} (MS-COCO) is utilized. In this work, 2017 version is used, which contains more than 100000 images from 79 classes. Since the most of the recent state-of-the-art object detectioon algorithms are evaluated by using MS-COCO, this dataset is selected for evaluation of the proposed method.

For the training, training set consisting more than 118000 images is utilized after applying the proposed image augmentation techniques. For the validation, validation set that contains more than 5000 images is utilized.

\section{Experimental Results}
We evaluated the prediction performance of our proposed method on both $test-dev-set$ of the VisDrone dataset \cite{pailla2019visdrone} and $val-2017$ set of the MS-COCO dataset \cite{lin2014microsoft}. For the evaluation metric, a common metric: mean average precision (mAP) is utilized. Average precision is defined as the area under the precision-recall curve. We computed three different average precision metrics: $mAP_{0.50}$, $mAP_{0.75}$ and $mAP_C$. For $mAP_{0.50}$, in order to consider a bounding box prediciton as true, intersection of union score (IoU) between the predicted and the ground truth bounding box must be higher than $0.50$. Similarly, for  $mAP_{0.75}$, bounding box predictions with IoU scores higher than $0.75$ are considered as true. Lastly, for $mAP_C$, average precisions are calculated for different IoU scores varying in the range of: $0.50:0.05:0.95$, and then, the mean value of those computed average precisions is used. 

\begin{table}
\centering
\caption{Results on $test-set$ of VisDrone \cite{pailla2019visdrone} dataset.}
\begin{tabular}{|l|c|c|c|} 
\hline
              & $mAP_{C}$ & $mAP_{0.50}$ & $mAP_{0.75}$  \\ 
\hhline{|====|}
SyNet (ours)  & \textbf{25.1}      & \textbf{48.4}        & \textbf{26.2}          \\ 
\hline
Cascade R-CNN & 24.7      & 43.7         & 24.3          \\ 
\hline
CenterNet     & 14.3      & 26.6         & 13.1          \\
\hline
\end{tabular}
\end{table}

\begin{table}
\centering
\caption{Results on individual classes using VisDrone \cite{pailla2019visdrone} $test-dev-set$ (for $mAP_{0.50}$).}
\label{table:VisDrone_results}
\begin{tabular}{|l|c|c|c|} 
\hline
           & SyNet (ours)    & Cascade R-CNN & CenterNet  \\ 
\hhline{|====|}
Pedestrian & \textbf{48.1}   & 42.6          & 22.6       \\ 
\hline
People     & \textbf{37.8}   & 33.1          & 20.6       \\ 
\hline
Bicycle    & \textbf{23.8}   & 21.2          & 14.6       \\ 
\hline
Car        & \textbf{83.2}   & 79.8          & 59.7       \\ 
\hline
Van        & \textbf{55.4}  & 49.3          & 24.0       \\ 
\hline
Truck      & \textbf{49.3}  & 43.5          & 21.3       \\ 
\hline
Tricycle   & \textbf{34.2}   & 31.6          & 20.1       \\ 
\hline
Awning     & \textbf{24.2}   & 21.5          & 17.4       \\ 
\hline
Bus        & \textbf{66.0}   & 61.9          & 37.9       \\ 
\hline
Motor      & \textbf{44.8}   & 43.1          & 23.7       \\
\hline
\end{tabular}
\end{table}

\ifx true false
\begin{table}
\centering
\caption{Results on $test-dev-set$ of VisDrone \cite{pailla2019visdrone} dataset.}
\begin{tabular}{|l|c|c|c|} 
\hline
              & $mAP_{C}$      & $mAP_{0.50}$   & $mAP_{0.75}$    \\ 
\hhline{|====|}
SyNet (ours)  & \textbf{29.1} & 52.4 & \textbf{28.6}  \\ 
\hline
RRNet \cite{RRNet}         & \textbf{29.1}          & \textbf{55.8}          & 27.2           \\ 
\hline
CornerNet \cite{pailla2019visdrone}     & 17.41          & 34.12          & 15.78           \\ 
\hline
Light-R-CNN \cite{pailla2019visdrone}   & 16.53          & 32.78          & 15.13           \\ 
\hline
FPN \cite{pailla2019visdrone}           & 16.51          & 32.20          & 14.91           \\ 
\hline
Cascade R-CNN \cite{pailla2019visdrone} & 16.09          & 31.91          & 15.01           \\ 
\hline
DetNet59 \cite{pailla2019visdrone}      & 15.26          & 29.23          & 14.34           \\ 
\hline
RefineDet \cite{pailla2019visdrone}     & 14.90          & 28.76          & 14.08           \\ 
\hline
RetinaNet \cite{pailla2019visdrone}     & 11.81          & 21.37          & 11.62           \\ \hline
\label{table:VisDrone_results}
\end{tabular}
\end{table}
\fi

Results for VisDrone dataset are presented in Table 1. To obtain Cascade R-CNN and CenterNet results on VisDrone dataset, those methods are trained on augmented VisDrone dataset. As this table shows, SyNet may be considered as state-of-the-art object detection method for aerial object detection since the proposed approach demonstrates a better performance for $mAP_{C}$, $mAP_{0.50}$ and $mAP_{0.75}$ than the recent state-of-the-art methods, Cascade R-CNN and CenterNet. Furthermore, in order to present the performance of the proposed network, average precision scores are presented for each class in VisDrone dataset in Table 2, which also shows that SyNet performs better than Cascade R-CNN and CenterNet. %However, RRNet \cite{RRNet} performed significantly better than our approach for $mAP_{0:50}$ metric. This difference may cause from the extensive image augmentation method utilized in RRNet, AdaResampling \cite{RRNet}.

\begin{table}
\centering
\caption{Results on $val-2017$ set of MS-COCO \cite{lin2014microsoft} dataset.}
\label{table:COCO_Results}
\begin{tabular}{|l|l|c|c|c|} 
\hline
\multicolumn{1}{|c|}{ \textbf{} }                                                                               & \multicolumn{1}{c|}{Backbone}                                 & $mAP_C$        & $mAP_{50}$     & $mAP_{75}$      \\ 
\hhline{|=====|}
SyNet (ours)                                                                                                    & \begin{tabular}[c]{@{}l@{}}ResNet101 \\+ DLA-34 \end{tabular} & \textbf{47.2}  & \textbf{66.4}  & \textbf{52.1}   \\ 
\hline
\multirow{2}{*}{\begin{tabular}[c]{@{}l@{}}Cascade \\R-CNN \end{tabular}}                                       & ResNet101                                                     & 42.7           & 61.6           & 46.6            \\ 
\cline{2-5}
                                                                                                                & ResNet50                                                      & 40.3           & 59.4           & 43.7            \\ 
\hline
\multirow{2}{*}{CenterNet}                                                                                      & Hourglass-104                                                 & 40.3           & 59.1           & 44.0            \\ 
\cline{2-5}
                                                                                                                & DLA-34                                                        & 37.4           & 55.1           & 40.8            \\ 
\hline
\multirow{2}{*}{\begin{tabular}[c]{@{}l@{}}Faster \\R-CNN \end{tabular}}                                        & ResNet101                                                     & 38.5           & 60.3           & 41.6            \\ 
\cline{2-5}
                                                                                                                & ResNet50                                                      & 36.4           & 58.4           & 39.1            \\ 
\hline
\multirow{2}{*}{\begin{tabular}[c]{@{}l@{}}Mask \\R-CNN \end{tabular}}                                          & ResNet101                                                     & 39.4           & 60.9           & 43.3            \\ 
\cline{2-5}
                                                                                                                & ResNet50                                                      & 37.3           & 59.0           & 40.2            \\ 
\hline
\multirow{2}{*}{Retina Net}                                                                                     & ResNet101                                                     & 37.7           & 57.5           & 40.4            \\ 
\cline{2-5}
                                                                                                                & ResNet50                                                      & 35.6           & 55.5           & 38.3            \\ 
\hline
\multirow{2}{*}{\begin{tabular}[c]{@{}l@{}}Cascade\\Mask R-CNN\textasciitilde{}\textasciitilde{} \end{tabular}} & ResNet101                                                     & 42.6           & 60.7           & 46.7            \\ 
\cline{2-5}
                                                                                                                & ResNet50                                                      & 41.2           & 59.1           & 45.1            \\ 
\hline
\multirow{2}{*}{\begin{tabular}[c]{@{}l@{}}Hybrid Task\\Cascade \end{tabular}}                                  & ResNet101                                                     & 44.9           & 63.8           & 48.7            \\ 
\cline{2-5}
                                                                                                                & ResNet50                                                      & 43.2           & 62.1           & 46.8            \\
\hline
\end{tabular}
\end{table}

Results for MS-COCO dataset are presented in Table 2. Results for other methods  are taken from \cite{cai2019cascadercnn}, \cite{chen2019mmdetection} and \cite{zhou2019objects}. SyNet achieved the best performance in the literature and out-performed many state-of-the-art methods for $mAP_{c}$, $mAP_{0.50}$ and $mAP_{0.75}$ metrics. However, a version of the EfficientDet \cite{tan2019efficientdet}, referred as EfficientDet-D7 (1536) achieved better result on $mAP_{c}$ metric, whose $mAP_{0.50}$ and $mAP_{0.75}$ results on validation set are not calculated thus omitted in the table. As result, SyNet not only performs well in aerial images, but performs well in both aerial images and ground images. 

\begin{figure*} 
    \centering
    \subfloat[]{%
       \includegraphics[width=0.24\linewidth]{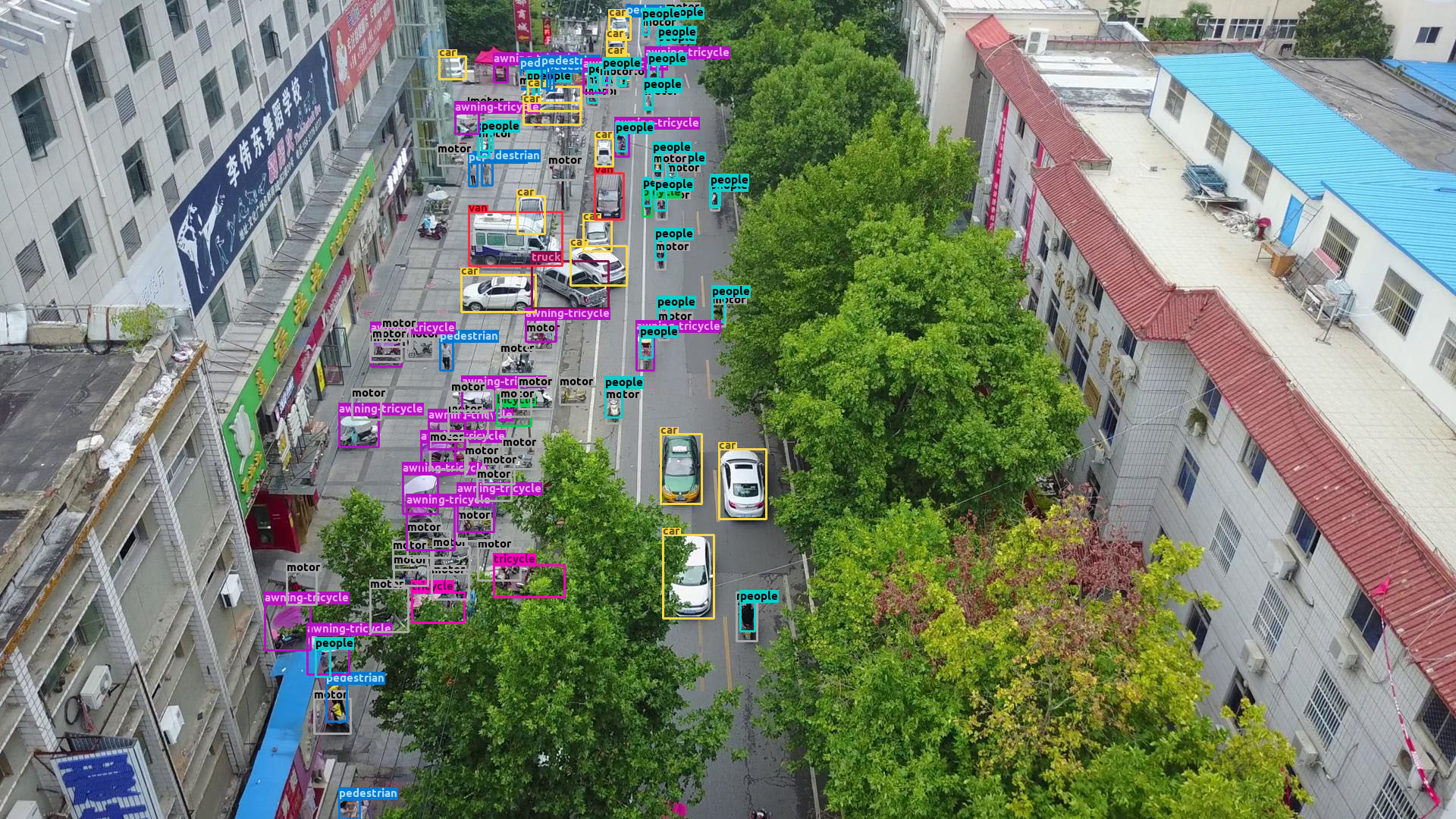}}
    \hfill
    \subfloat[]{%
       \includegraphics[width=0.24\linewidth]{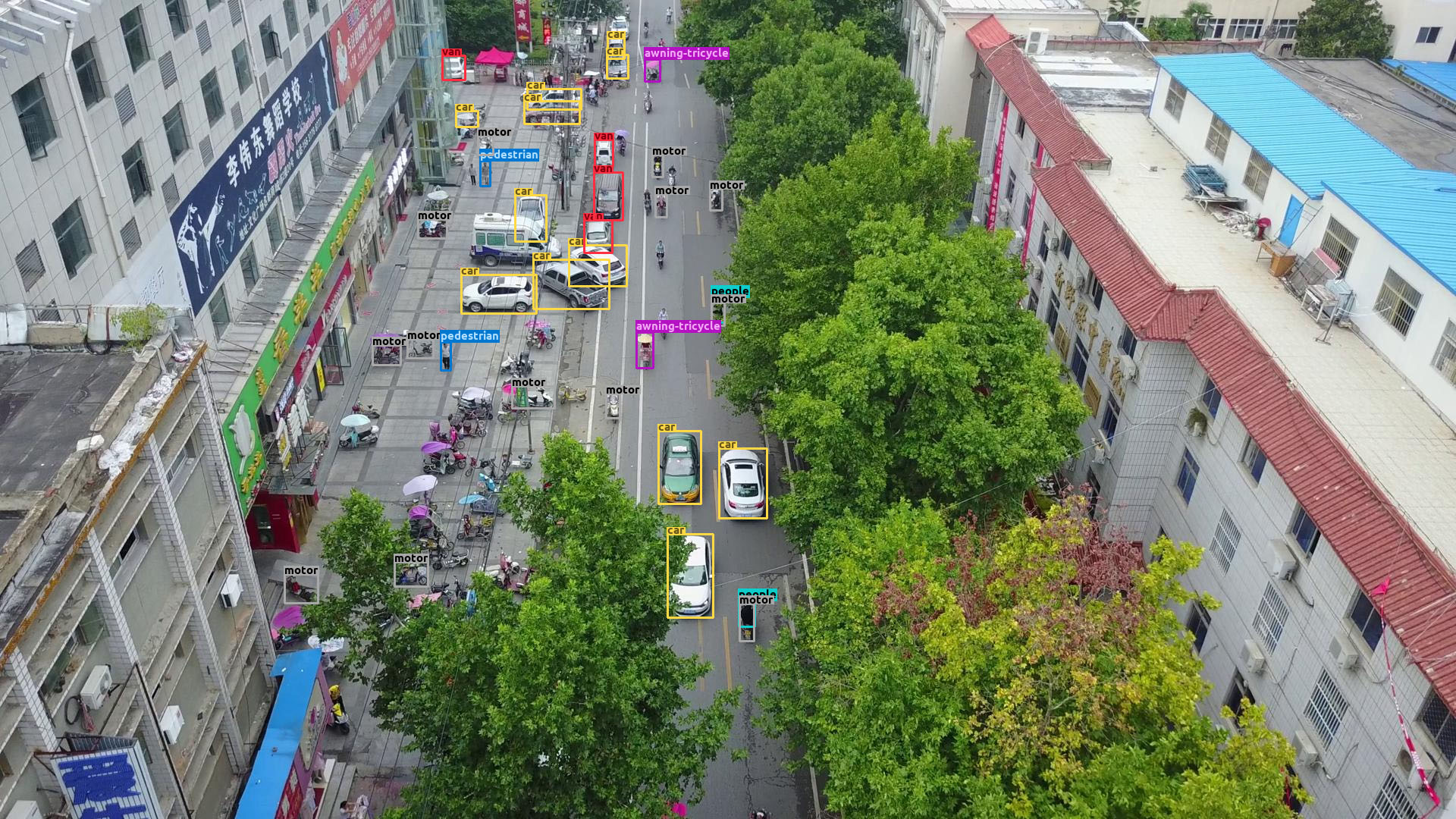}}
    \hfill
    \subfloat[]{%
       \includegraphics[width=0.24\linewidth]{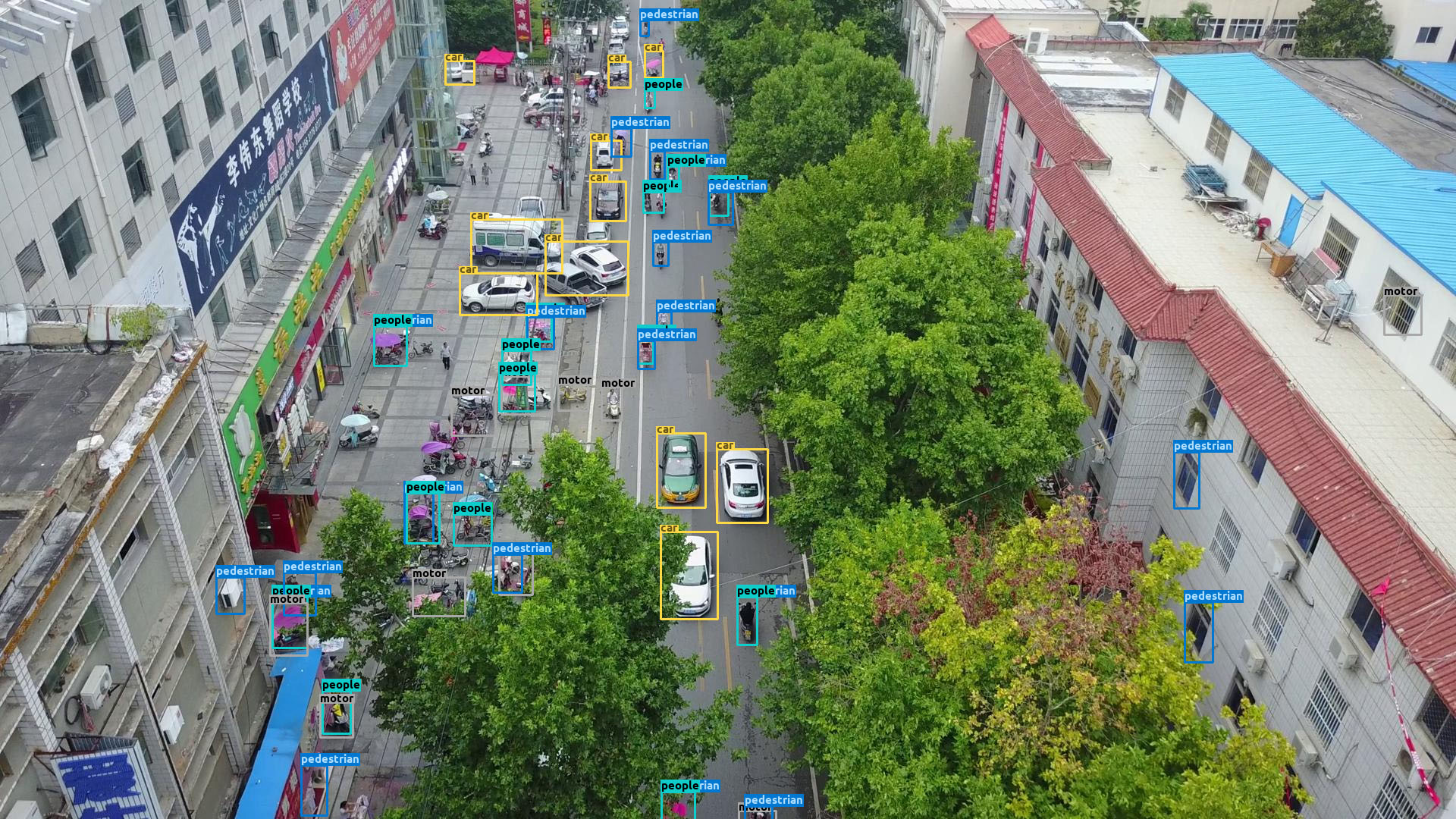}}
    \hfill
    \subfloat[]{%
        \includegraphics[width=0.24\linewidth]{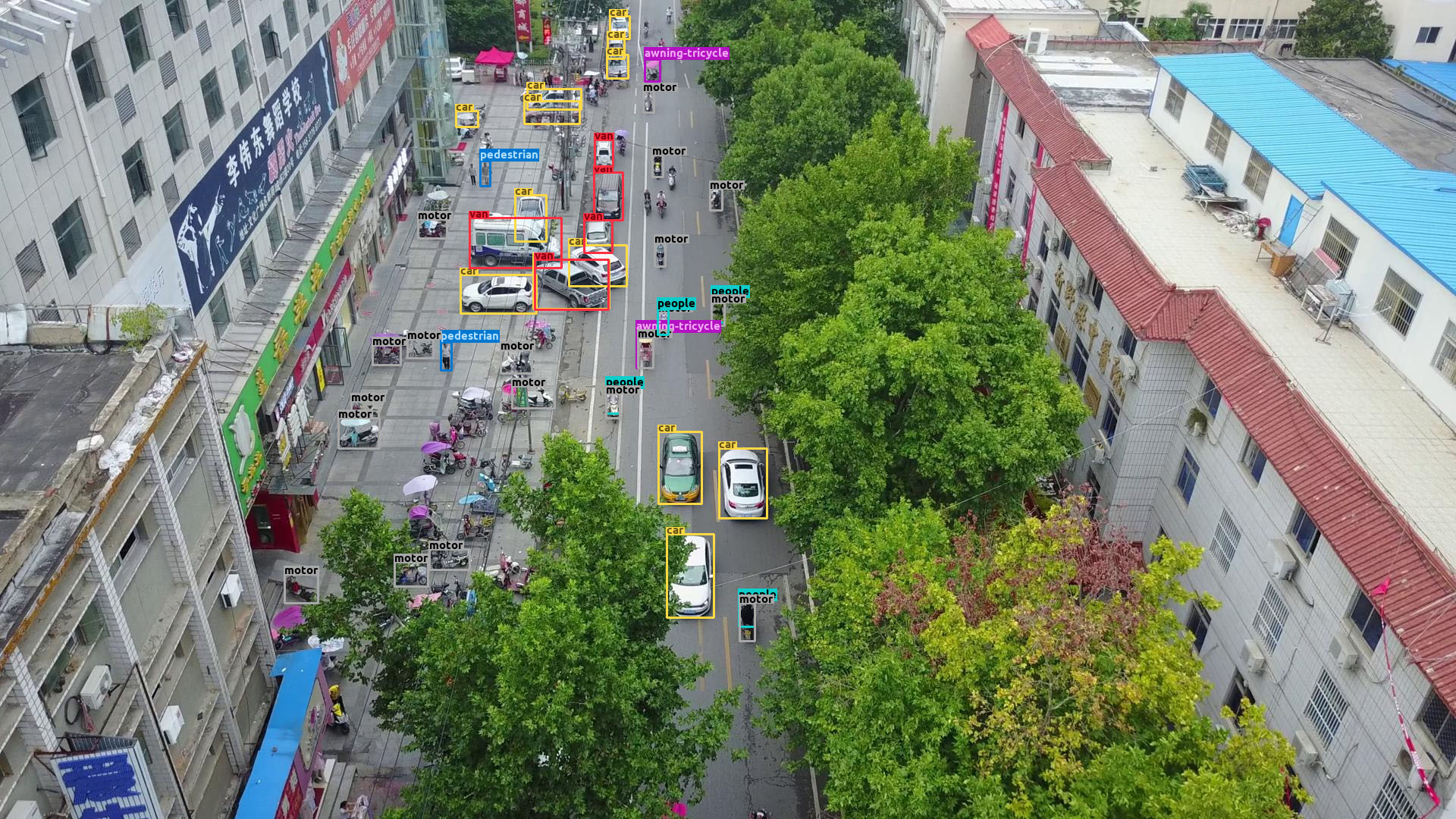}}
    \hfill
    \subfloat[]{%
       \includegraphics[width=0.24\linewidth]{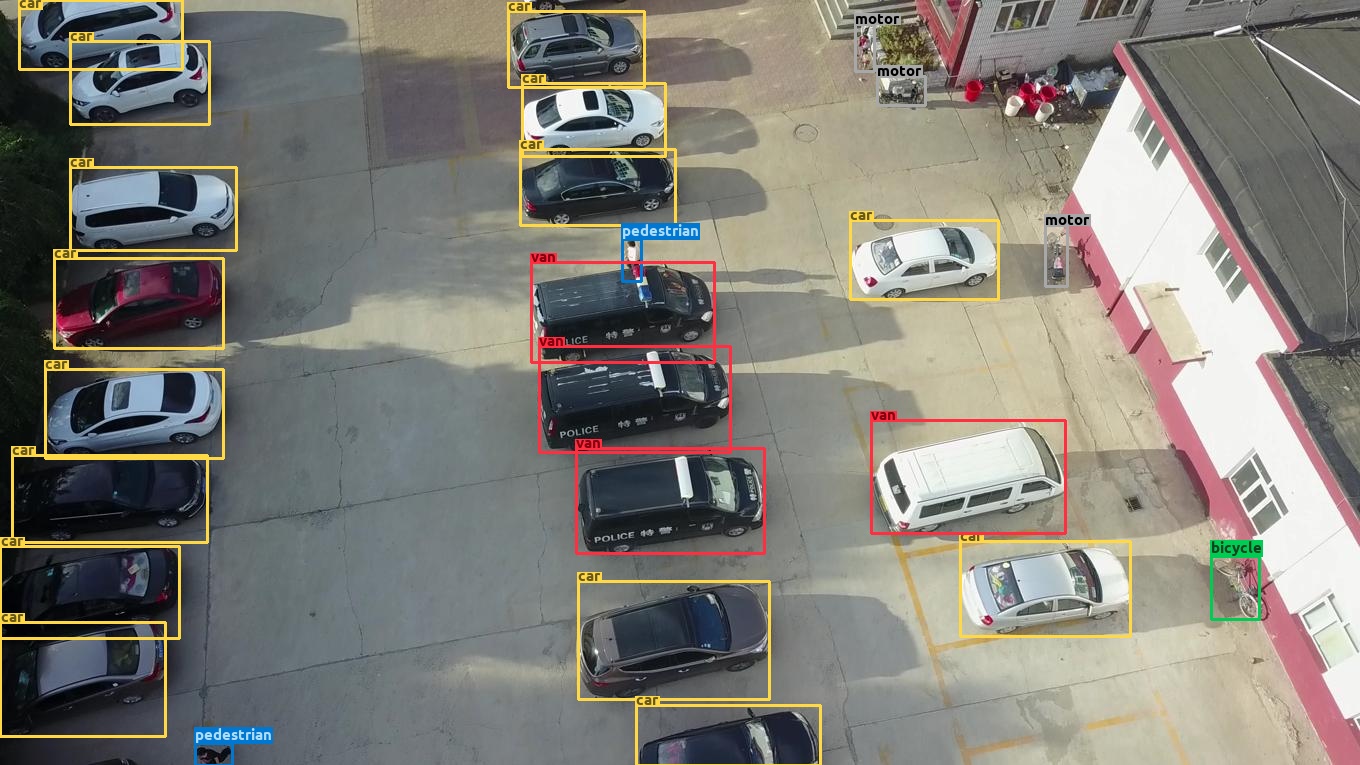}}
    \hfill
    \subfloat[]{%
       \includegraphics[width=0.24\linewidth]{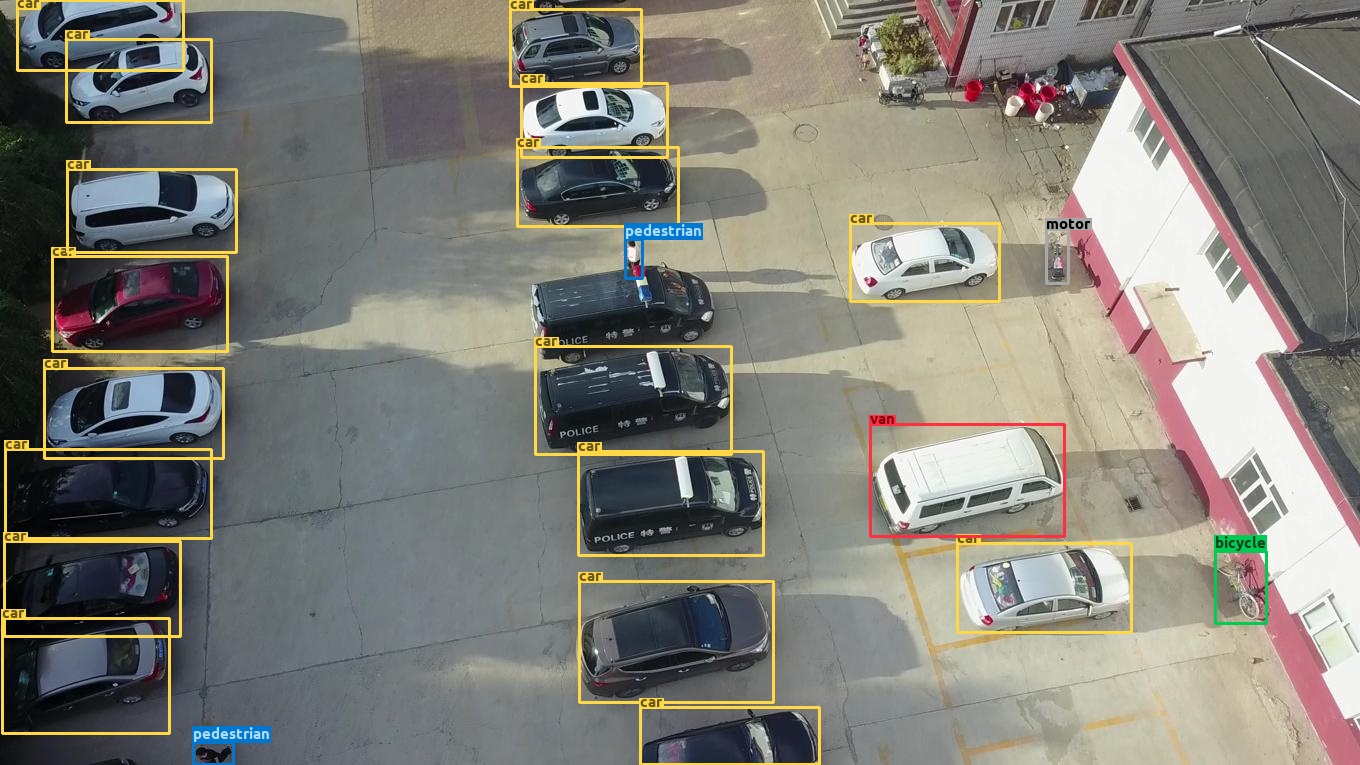}}
    \hfill
    \subfloat[]{%
       \includegraphics[width=0.24\linewidth]{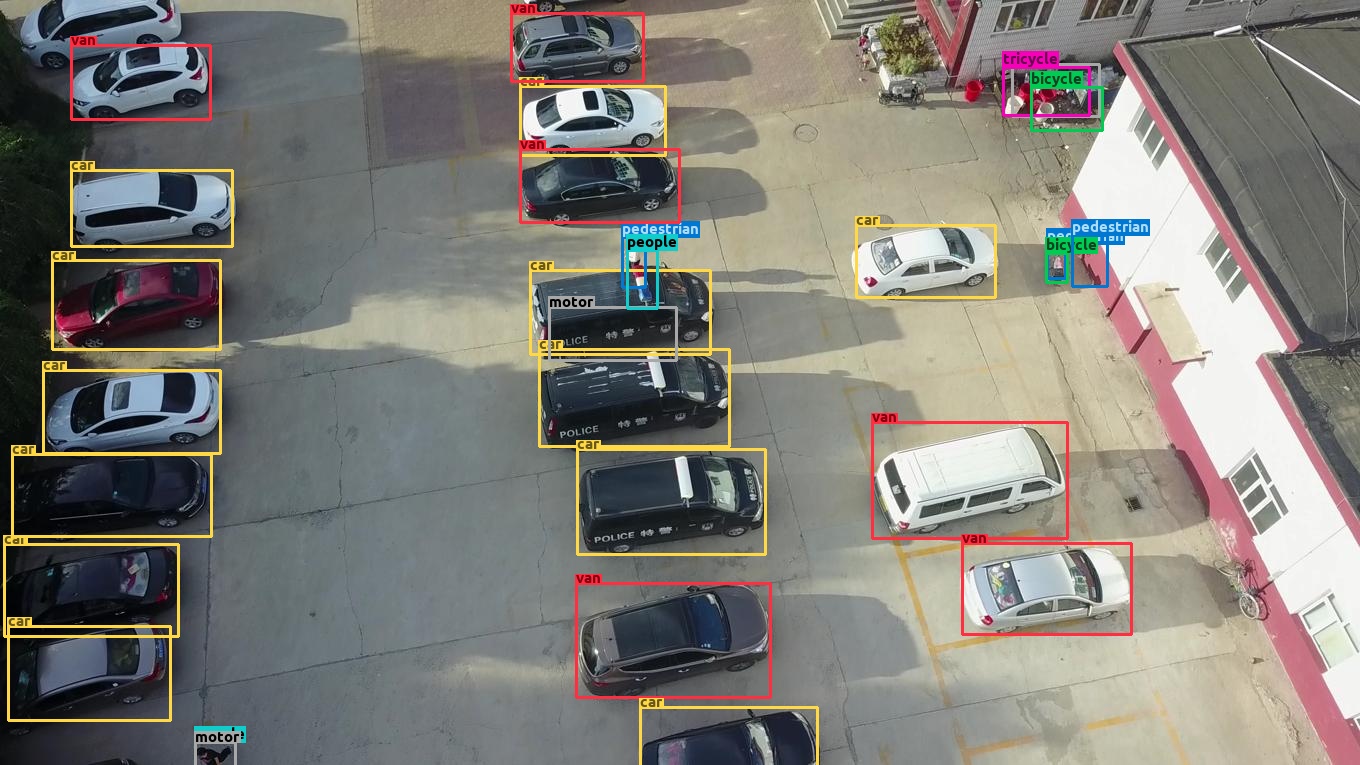}}
    \hfill
  \subfloat[]{%
        \includegraphics[width=0.24\linewidth]{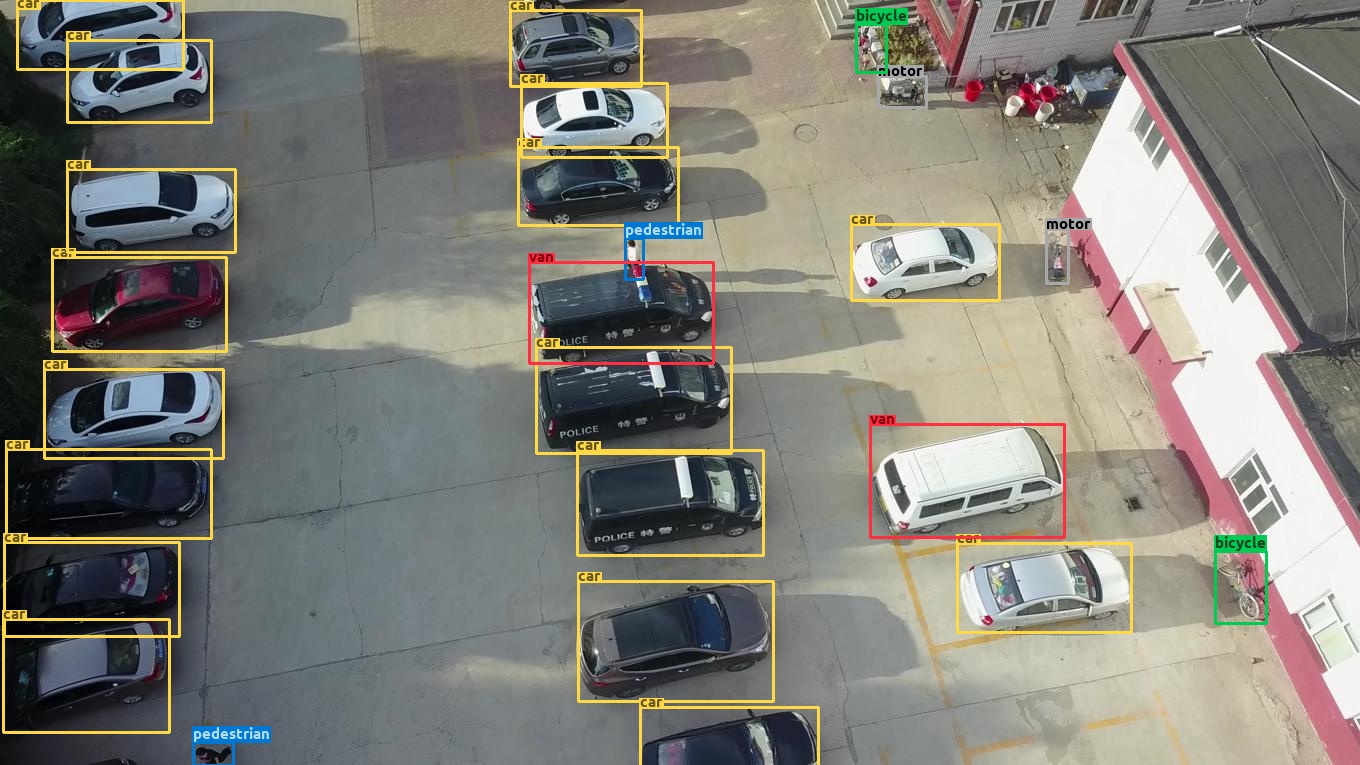}}
  \caption{Results on two sample images, in which only bounding boxes with more than 0.5 confidence score are shown: (a) Ground truth 1, (b) Cascade R-CNN result 1, (c) CenterNet result 1, (d) SyNet result 1, (e) Ground truth 2, (f) Cascade R-CNN result 2, (g) CenterNet result 2 and (h) SyNet result 2.}
  \label{augm1} 
\end{figure*}

Besides showing some quantitative results, we also demonstrate some qualitative results as shown in Figure~\ref{augm1}. For instance, second example shows that there are four vans in the image, Cascade R-CNN and CenterNet predicted only one of them correctly but SyNet predicted two of them correctly. % and in Figure~\ref{augm2}.

%	\vspace{-0.4in}

\section{Conclusion}

In this paper, we introduced a novel ensemble network for object detection yielding state of the art results when compared to the existing techniques on drone images. Our proposed method, SyNet, combines a multi-stage detector and a single-stage detector and makes a prediction by combining the individual predictions of each algorithm through a fusion stage. By carefully choosing the the individual algorithms, we made use of our their advantageous properties in detection and combined them in our emsemble network. As demonstrated in our experiments, our proposed approach yields the highest $mAP_C$ and $mAP_{0.75}$ scores in both datasets: VisDrone and in MS-COCO, when compared to most recently proposed state of the art detection algorithms.  

While our SyNet performs better in Table~\ref{table:VisDrone_results} and yields higher results than Cascade R-CNN and CenterNet solutions in all classes on VisDrone dataset, as the results of both Table~\ref{table:COCO_Results} and Table~\ref{table:VisDrone_results} suggests, the average  $mAP_{0.50}$  value in Table~\ref{table:COCO_Results} remain significantly lower than the average $mAP_{0.50}$ value of Table~\ref{table:VisDrone_results}. That analysis suggests that the object detection task with the existing state of the art object detection algorithms still remains as a more challenging problem in drone images, when compared to detecting objects in ground taken images. 

% Future work may include developing larger ensembling strategies and developing fusion schemes that combines the individual decisions in a more effective way.

\section*{Acknowledgment}
\small{This paper has been produced benefiting from the 2232 International Fellowship for Outstanding Researchers Program of TÜBİTAK (Project No:118C356). However, the entire responsibility of the paper belongs to the owner of the paper. The financial support received from TÜBİTAK does not mean that the content of the publication is approved in a scientific sense by TÜBİTAK.}

%to obtain a high quality, state-of-the-art object detector. As building blocks of the combination network, SyNet, Cascade R-CNN and CenterNet are utilized. Furthermore, image augmentation techniques are employed along with ensembling functions.

%Proposed approach is validated by using two different datasets: 1. MS-COCO $val-2017$ set and 2. VisDrone $test-set$. As result, it is obtained that proposed approach perform better than most of the state-of-the-art methods in the literature.

%As future work, a better augmentation approach may be employed to achieve better performance. Furthermore, different assembling approaches may be utilized.

\bibliographystyle{IEEEtran}\bibliography{biblo}

\end{document}